\documentclass[10pt,twocolumn,letterpaper]{article}

\usepackage{cvpr}              

\usepackage{graphicx}
\usepackage{xcolor}
\usepackage{colortbl}
\usepackage[export]{adjustbox}

\usepackage{wrapfig}
\usepackage{multirow}
\usepackage{multicol}
\usepackage{makecell}
\usepackage{arydshln}
\usepackage{subcaption}
\usepackage{tabularx}

\definecolor{navyblue}{HTML}{0071BC}

\newcommand{\ci}[3]{\ensuremath{#1\%\;{}^{+#2}_{-#3}}}
\newcolumntype{N}{>{\centering\arraybackslash\LARGE}c}
\newcolumntype{L}{>{\raggedright\arraybackslash}X}
\newcolumntype{P}[1]{>{\raggedright\arraybackslash}p{#1}}

\definecolor{gpt4turbo}{HTML}{b0be90}
\definecolor{gpt4o}{HTML}{545d3e}
\definecolor{claude35}{HTML}{EF7C24}
\definecolor{claude3}{HTML}{ffd9b3}
\definecolor{llama}{HTML}{bfd9ff}
\definecolor{qwen2b}{HTML}{e0cab3}
\definecolor{qwen4b}{HTML}{b6834e}
\definecolor{qwen8b}{HTML}{a1601b}
\definecolor{bestclosed}{HTML}{FFD6D6} 
\definecolor{bestopen}{HTML}{FFADAD}  
         
\newif\ifdoublerule

\definecolor{cvprblue}{rgb}{0.21,0.49,0.74}
\usepackage[pagebackref,breaklinks,colorlinks,allcolors=cvprblue]{hyperref}

\def\paperID{xxxx}
\def\confName{CVPR}
\def\confYear{2026}

\definecolor{wrm-blue}{HTML}{7FB3FF} 

\hypersetup{
    colorlinks=true,
    linkcolor=wrm-blue,
    filecolor=wrm-blue,
    urlcolor=wrm-blue,
    citecolor=wrm-blue,
}

\newcommand{\metricLabel}[1]{prediction correctness}
\newcommand{\metricLabelb}[1]{Prediction correctness}
\newcommand{\metricLabelcp}[1]{Prediction correctnesses}
\newcommand{\metricLabelcapital}[1]{Prediction Correctnesses}
\title{Beyond Recognition: Evaluating Visual Perspective Taking \\ in Vision Language Models}

\author{
\begin{tabular}{ccc}
Gracjan Góral$^{1,2,6*}$ & Alicja Ziarko$^{1,2}$ & Piotr Miłoś$^{1,2}$ \\
Michał Nauman$^{1,4}$ & Maciej Wołczyk$^{5}$ & Michał Kosiński$^{3}$
\end{tabular}
\\
\\
\begin{tabular}{c}
$^{1}$University of Warsaw \quad
$^{2}$Polish Academy of Sciences \quad
$^{3}$Stanford University\\
$^{4}$University of California, Berkeley \quad
$^{5}$ IDEAS NCBR \quad
$^{6}$Lute\\[6pt]
$^{*}$Corresponding author: {\tt gp.goral@uw.edu.pl}
\end{tabular}
}

\begin{document}
\maketitle
\begin{abstract}
We investigate the ability of Vision Language Models (VLMs) to perform visual perspective taking using a new set of visual tasks inspired by established human tests. Our approach leverages carefully controlled scenes in which a single humanoid minifigure is paired with a single object. By systematically varying spatial configurations—such as object position relative to the minifigure and the minifigure’s orientation—and using both bird’s-eye and surface-level views, we created 144 unique visual tasks. Each task is paired with a series of $7$ diagnostic questions designed to assess three levels of visual cognition: \textit{scene understanding, spatial reasoning, and visual perspective taking}.
We evaluate several high-performing models, including Gemini Robotics-ER 1.5, Llama-3.2-11B-Vision-Instruct, and variants of Claude Sonnet, GPT-4, and Qwen3, and find that while they excel at scene understanding, performance declines markedly on spatial reasoning and deteriorates further on perspective taking. Our analysis suggests a gap between surface-level object recognition and the deeper spatial and perspective reasoning required for complex visual tasks, pointing to the need for integrating explicit geometric representations and tailored training protocols in future VLM development.
\end{abstract}

\section{Introduction}
\begin{figure*}[ht!]
    \centering
    \includegraphics[width=\textwidth]{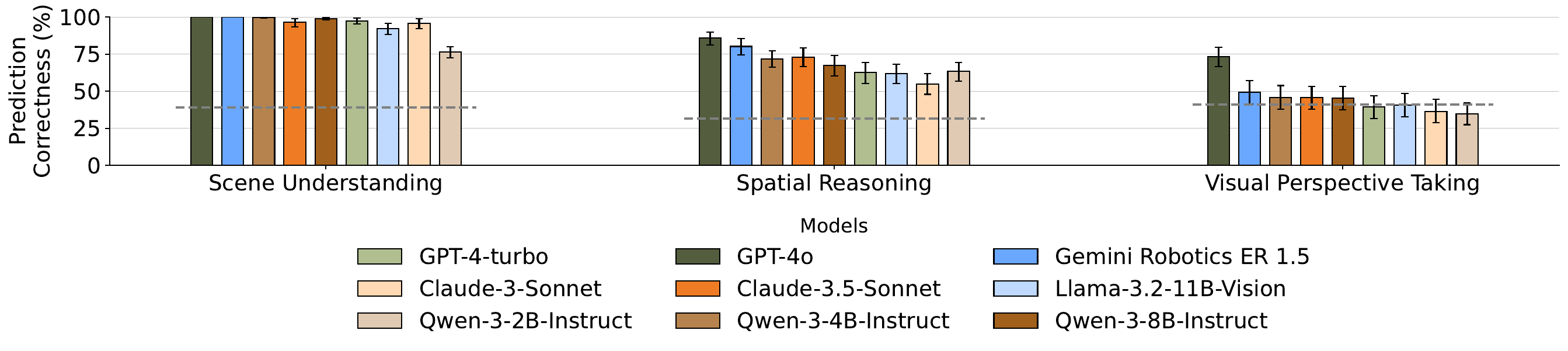}
    \caption{\small \metricLabelb{} across three categories of growing difficulty: \emph{scene understanding}, \emph{spatial reasoning}, and \emph{visual perspective taking}. Error bars represent $95\%$ confidence intervals (estimated using bootstrapping (10,000 iterations)). The random classifier is a baseline choosing an answer uniformly at random, see \ref{app:random_baseline}.}
    \label{fig:main_fig}
\end{figure*}

Recent advances in vision–language modeling have produced systems capable of jointly reasoning over images and text~\cite{bubeck2023sparksartificialgeneralintelligence, aghajanyan2023scaling, bai2023qwenvlversatilevisionlanguagemodel}. These models are being applied across diverse domains, including robotics and autonomous driving~\cite{ding2024quarvlavisionlanguageactionmodelquadruped, sathyamoorthy2024convoicontextawarenavigationusing}, as well as healthcare~\cite{hartsock2024vision, ferber2024incontext, tanno2024collaboration}. However, excelling in such applications requires the ability to reason about what others can and cannot see, going beyond traditional scene recognition or language understanding. For instance, an autonomous vehicle must anticipate what a nearby driver can or cannot see, while a surgical or industrial robot should assess whether a human coworker can visually locate an object before passing it. Reasoning about visual perspectives is therefore essential for safe, collaborative, and socially aware embodied systems.

In humans, the ability to adopt another’s visual vantage point, known as visual perspective taking (VPT)~\cite{Piaget1956, Thorsten, Lukosiunaite2024}, is a core component of theory of mind~\cite{heyes2014cultural}. VPT supports a wide range of cognitive and social functions, from spatial navigation to joint action coordination, and impairments in VPT have been linked to difficulties in both domains~\cite{Orefice2024-oh, pearson2013review}. Many cognitive abilities that have been central to the development of machine learning, such as perception and reasoning, were first characterized in humans, with their testing paradigms later adapted to evaluate machines. Because VPT has been extensively studied in psychology with well-established tasks, it provides a natural foundation for assessing similar abilities in foundation models.

Recent studies suggest that large language models can track the beliefs of different characters, effectively reasoning about the world \textit{through the eyes of their minds} in the linguistic domain~\cite{kosinski2024evaluating, gandhi2023understandingsocialreasoninglanguage, Strachan2024}. This raises the question of whether similar perspective-taking abilities extend to the visual domain. To address this, we build on the rich psychological literature on human VPT~\cite{Loomis2003, Kessler2014} and introduce  Isle-Bric-V2, a controlled dataset of 144 manually created visual tasks by systematically varying spatial configurations of humanoid minifigures and objects, with each image accompanied by 7 questions testing scene understanding, spatial reasoning, and visual perspective taking. We use this dataset to investigate whether VLMs can take others' visual perspectives—that is, whether they can see the world through another's eyes

Unlike large-scale benchmarks that prioritize breadth and are susceptible to data contamination~\cite{balloccu-etal-2024-leak}, our approach emphasizes experimental control and variable isolation. We adopt a \textit{minimal-contrast} methodology inspired by cognitive science: conditions are constructed so that they differ in only a single cognitively relevant factor (e.g., the humanoid minifigure's orientation or the camera viewpoint), while all other aspects of the scene are held constant. This is analogous to classic dot-perspective paradigms in psychology, where otherwise identical displays differ only in what an avatar can see~\cite{ogrady2020perspective,RUBIOFERNANDEZ2022108256}, and to cognition-inspired benchmarks for language models such as COMPS, which use conceptual minimal pairs to probe structured knowledge and world modeling~\cite{misra-etal-2023-comps}.

\paragraph{Contributions}
Our work makes three key contributions:

\textbf{(1)} We introduce Isle-Brick-V2, a manually created, psychologically grounded diagnostic benchmark for visual perspective taking that uses open-ended questions to cleanly separate scene recognition, spatial reasoning, and perspective taking abilities on the same set of stimuli.

\textbf{(2)} We systematically evaluate nine recent VLMs on this benchmark, revealing consistent VPT failures and a persistent directional prior that remains robust under multiple visual and text interventions.

\textbf{(3)} We present empirical evidence that accurate orientation detection does not guarantee correct perspective-taking. Our findings are consistent with a partial dissociation between orientation detection and perspective-taking in our setting, motivating future architectures and training protocols that incorporate explicit egocentric or geometric representations.

\section{Contributions in Context of Related Work}
\begin{figure*}[ht!]
  \centering
  \includegraphics[width=0.8\linewidth]{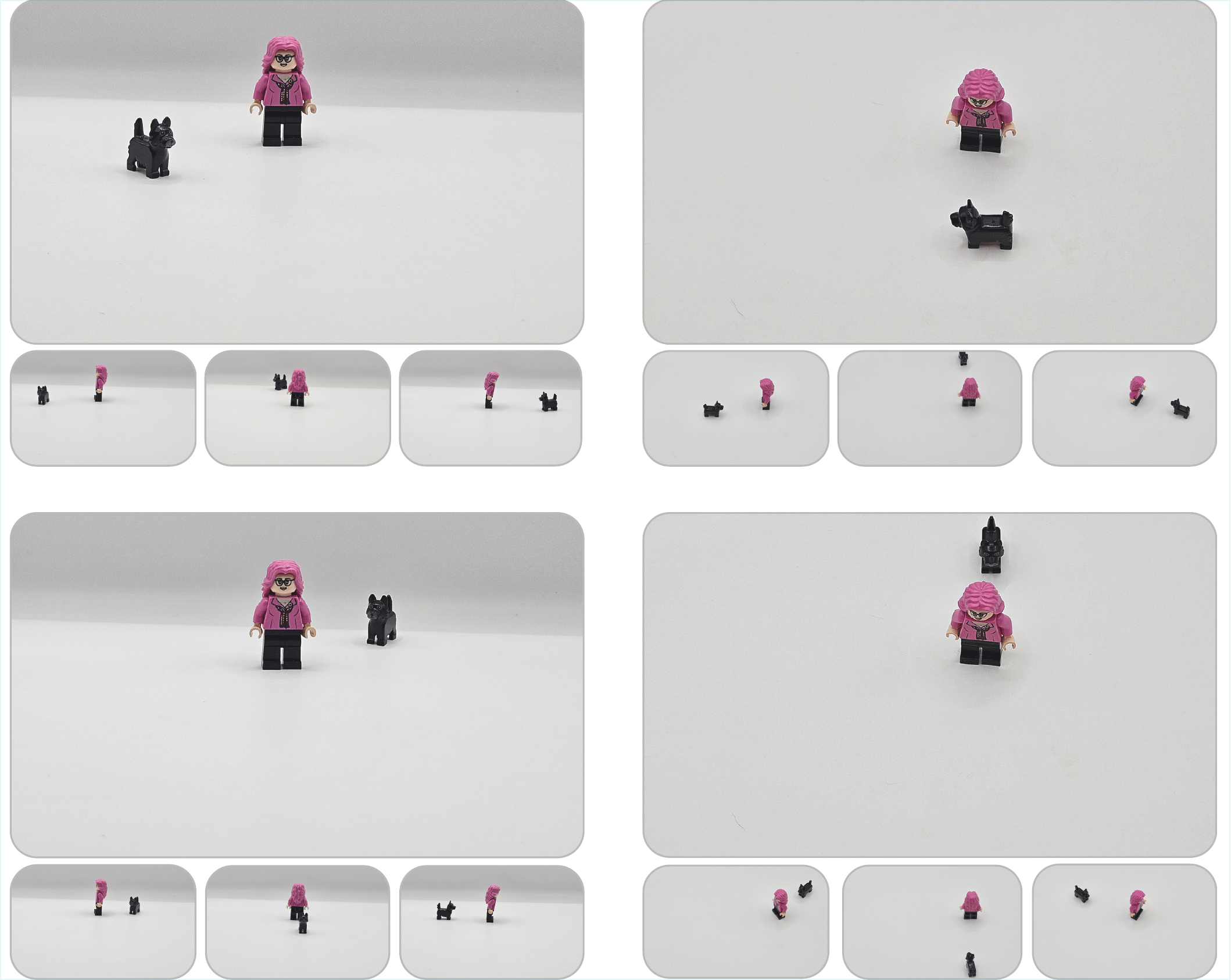} 
  \caption{\small Sixteen tasks involving a single humanoid minifigure--object pair. Tasks vary by the object's placement (left, right, front, back); the orientation of the humanoid minifigure (facing toward or away from the object); and camera angle (surface-level and bird's-eye views). All images had the same dimensions, but some are enlarged here for presentation purposes. \vspace{-5pt}}
\label{fig:dataset_example}
\end{figure*}
\paragraph{Psychological foundations.}
Within the psychological literature, VPT is often conceptualized along two hierarchically organized levels~\cite{Flavell1981Young,Moll2006Level1P,samson2010seeing}. Level~1 concerns understanding what others can or cannot see (e.g., \emph{Do they see the object?}), whereas Level~2 involves mentally adopting another’s vantage point to determine how objects appear from that perspective (e.g., \emph{From their viewpoint, is the object to the right or left?}). In adults, Level-1 judgments are typically supported by rapid line-of-sight computations~\citep{Michelon2006TwoKinds,Kessler2010TwoForms}, whereas Level-2 judgments require more effortful mental transformations and are associated with mental rotation costs~\cite{Surtees2013Similarities,janczyk2013}. The mere presence of others can trigger spontaneous perspective changes~\citep{Tversky2009Embodied}, although the degree of automaticity in VPT remains debated~\citep{Surtees2013Similarities}.

\textit{Our contribution.}
We explicitly ground our diagnostic tests in the Level-1 / Level-2 framework to separate qualitatively different forms of VPT failure. The same visual stimuli are probed with questions targeting visibility (Level~1) and viewpoint-dependent spatial relations (Level~2), yielding a testing procedure that is in principle applicable to both human participants and VLMs.

\vspace{-13pt}
\paragraph{Spatial reasoning and VPT in VLMs.}
Recent VLMs such as GPT-4, Gemini, and Claude report strong performance on a variety of visual and spatial benchmarks probing object relations, counting, and compositional reasoning~\citep{openai2024gpt4technicalreport,geminiteam2024geminifamilyhighlycapable,Anthropic2024Claude3}. Specialized approaches like SpatialVLM and SpatialRGPT further improve spatial understanding by leveraging internet-scale 3D data, metric depth, and region-level grounding~\citep{Chen2024SpatialVLM,Cheng2024SpatialRGPT}, while ScanQA and SQA3D extend visual question answering to RGB-D and 3D environments~\citep{azuma_2022_CVPR,ma2022sqa3d}. The 3D-PC benchmark evaluates visual perspective taking via depth-ordering and line-of-sight classification in natural scenes~\citep{linsley20243d}, and Omni-Perspective scales multimodal ToM-style evaluation to thousands of question–context–answer tuples over real images and a multi-level hierarchy of social and mental-state reasoning~\citep{wang2025do}.

\textit{Our contribution.}
In contrast to 3D-PC and Omni-Perspective, which probe VPT in natural scenes via depth-order or multi-choice ToM-style queries~\citep{linsley20243d,wang2025do}, we introduce a psychology-grounded visual task benchmark. By asking a hierarchy of seven questions that range from scene recognition to visual perspective taking on the same stimuli, we can disentangle failures of basic object recognition from failures of spatial reasoning and perspective taking in a way that is difficult to achieve with existing large-scale 3D VPT benchmarks.

\section{Experimental Setup}

\subsection{Visual Tasks}
We developed 144 novel visual tasks inspired by the established human VPT tests~\cite{OGrady2020, lukosiunaite2024influence} to ensure the models had not encountered them during training and could not simply recall memorized solutions. The visual tasks represented a humanoid minifigure and an object placed on the same surface. We used nine distinct humanoid minifigures (varying in hairstyle, clothing, and accessories) and nine distinct objects: a plant, a wardrobe, a cat, a dog, a goblet, a chair, a desk, a bat, and a computer monitor. For each minifigure-object pair, we systematically varied their spatial positions, see Figure~\ref{fig:dataset_example}. These variations resulted in a total of 144 tasks (9 pairs × 4 spatial positions × 2 orientations × 2 viewpoints), which can be downloaded from~\url{[LINK]}.

The use of LEGO elements further enabled precise control over image and scene properties without post hoc manipulation. Thus, our approach complemented previous work relying on web-scraped images that may be part of training data~\cite{tafascatoward} or used abstract visual elements such as dots and arrows overlaid on real images~\cite{linsley20243d}.

\subsection{Diagnostic Questions}
\paragraph{Question Design.}
Each visual task was paired with a set of seven questions, see Table~\ref{tab:combined_questions}. 
Questions Q1, Q2, and Q3 focused on \textit{scene understanding}: Q1 asked for the total number of objects, Q2 asked for the number of humanoid minifigures, and Q3 inquired whether the humanoid minifigure and objects shared the same surface. Questions Q4 and Q5 tested \textit{spatial reasoning}: Q4 required identifying the object’s location relative to the humanoid minifigure, and Q5 queried the humanoid minifigure’s facing direction. Finally, Q6 and Q7 evaluated \textit{visual perspective taking}: Q6 checked if the humanoid minifigure could see the object (Level-1), while Q7 asked for the object’s location from the humanoid minifigure’s perspective (Level-2).

\vspace{-15pt}
\paragraph{Open-Ended Format.}
To reduce the influence of guessing and to avoid position-based artifacts common in multiple-choice evaluations~\cite{Pezeshkpour2023LargeLM}, all questions were presented in an open-ended format. Models were not constrained to choose from a fixed set of options and could produce free-form text, although we only scored the final answer. This setup matches typical user interactions and avoids hand-crafted, model-specific prompting strategies. Each question was answered independently in a zero-shot manner (with the context cleared between questions), using temperature~0 to minimize variance, and the maximum response length was capped at 128 tokens.

\begin{table*}[t]
\centering
\scriptsize
\renewcommand{\arraystretch}{1.5}
\setlength{\tabcolsep}{4pt}
\caption{\small List of the diagnostic questions with an example answer by GPT-4-Turbo for the task shown in the top-left panel of Figure~\ref{fig:dataset_example}.}
\label{tab:combined_questions}

\begin{tabularx}{\textwidth}{@{} P{0.43\textwidth} P{0.18\textwidth} L @{}}
\toprule
\multicolumn{1}{c}{\textbf{Question}} &
\multicolumn{1}{c}{\textbf{Ability}} &
\multicolumn{1}{c}{\textbf{GPT-4-Turbo’s answer}} \\
\midrule

\rowcolor{navyblue!5}
\multicolumn{3}{c}{\textbf{Scene understanding}} \\
\addlinespace[.3em]
Q1: \textit{List and count all objects in the image that are not humanoid minifigures.}
& Object recognition, \newline counting \cite{spelke2007core}
& \textit{In the image, there is one object that is not a humanoid minifigure: 1. A small black dog figure. So, there is a total of~1 non-humanoid object in the image.} \\[0.25em]

Q2: \textit{How many humanoid minifigures are in the picture?}
& Object recognition, \newline counting \cite{spelke2007core}
& \textit{There is one humanoid minifigure in the picture.} \\[0.25em]

Q3: \textit{Are the humanoid minifigure and the object on the same surface?}
& Object placement \cite{spelke1990principles}
& \textit{Yes, both the humanoid minifigure and the dog are on the same flat, white surface.} \\
\midrule

\rowcolor{navyblue!5}
\multicolumn{3}{c}{\textbf{Spatial reasoning}} \\
\addlinespace[.3em]
Q4: \textit{Assuming the top of the image is north, in which cardinal direction (i.e., north, west, east, or south) is the object located relative to the humanoid minifigure?}
& Spatial reasoning \cite{uttal2013malleability}
& \textit{The dog is located to the west of the humanoid minifigure, assuming the top of the image is north.} \\[0.25em]

Q5: \textit{Assuming the top of the image is north, which cardinal direction (i.e., north, west, east, or south) is the humanoid minifigure facing?}
& Mental rotation, spatial reasoning \cite{shepard1971mental}
& \textit{The humanoid minifigure is facing east.} \\
\midrule

\rowcolor{navyblue!5}
\multicolumn{3}{c}{\textbf{Visual perspective taking}} \\
\addlinespace[.3em]
Q6: \textit{Assuming that the humanoid minifigure can see and its eyes are open, does it see the object?}
& Understanding what \newline others see (Level-1) \cite{samson2010seeing}
& \textit{Yes, assuming the humanoid minifigure can see and its eyes are open, it appears to be facing the dog, so it would be able to see the dog.} \\[0.25em]

Q7: \textit{From the perspective of the humanoid minifigure, where is the object located relative to it? Please use terms such as front, left, right, or back.}
& Adopting others' \newline perspective (Level-2) \cite{surtees2012direct}
& \textit{From the perspective of the humanoid minifigure, the dog is located to its left side.} \\
\bottomrule
\end{tabularx}
\end{table*}

\paragraph{Gold-Standard Answers.}
To create the gold-standard answers for each of the 144 visual tasks paired with each of the 7 questions, three research assistants independently responded, with the initial agreement rate exceeding 99\%. The rare disagreements were resolved through discussion. Multiple gold-standard answers were allowed for certain questions, e.g., in Q5, where multiple cardinal directions could be correct (north and east for the humanoid minifigure facing northeast). The distribution of gold-standard answers for each question is presented in~\ref{app:gold_dist}.

\subsection{Evaluation Procedure} Two research assistants processed the model's responses, extracting the key answer components. This included details like object counts (Q1, Q2), spatial relationship confirmations (Q3), line-of-sight judgments (Q6), egocentric locations (Q7), and the cardinal directions (Q4, Q5). Rare disagreements, less than~1\%, were resolved by a third person. Combined answers (like northeast) were transformed into sets of basic directions (north, east). We evaluated performance using averaged \textit{\metricLabel{}}. This metric calculates precision for each answer: it is the fraction of components in the model's answer that are present in the gold-standard answer set. For example, if the model predicted \textit{northeast} (components: \textit{north}, \textit{east}) against a gold-standard answer of \textit{north}, the \metricLabel{} is one correct component divided by two predicted components, yielding 0.5. This approach quantifies partial correctness and reduces to standard accuracy for single-component answers. Further details are in Appendix \ref{app:acc}.
\section{Results}

\begin{table*}[ht!]
  \centering
  \small
\caption{
Each question was evaluated across 144 visual tasks in three categories: \textit{Scene Understanding} (random baseline: 38.9\%), \textit{Spatial Reasoning} (random baseline: 31.7\%), and \textit{Visual Perspective Taking} (random baseline: 41.1\%). \raisebox{-0.2ex}{\colorbox{bestclosed}{\rule{0pt}{1.2ex}\hspace{1em}}} indicates the best closed model performance, while \raisebox{-0.2ex}{\colorbox{bestopen}{\rule{0pt}{1.2ex}\hspace{1em}}} indicates the best open-source model performance. For VPT tasks, the best closed model (GPT-4o) achieves +32.15pp above random, while the best open-source model (Qwen3-4B-Instruct) achieves +4.75pp above random.}
  \label{tab:model_comparison}
  {\renewcommand{\arraystretch}{3}%
   \setlength{\tabcolsep}{7pt}%
   \begin{adjustbox}{max width=\textwidth,center}
   \begin{tabular}{l N N N N N N N N N}
    \toprule
    \ifdoublerule \specialrule{\heavyrulewidth}{0pt}{0pt}\fi
    \textbf{\Large Question} &
    \cellcolor{gpt4turbo!50}\textbf{\Large GPT-4 Turbo} &
    \cellcolor{gpt4o!50}\textbf{\Large GPT-4o} &
    \cellcolor{lime!10}\makecell{\textbf{\Large Gemini Robotics}\\\textbf{\Large ER 1.5}} &
    \cellcolor{claude3!50}\textbf{\Large Claude 3 Sonnet} &
    \cellcolor{claude35!50}\textbf{\Large Claude 3.5 Sonnet} &
    \cellcolor{llama!50}\makecell{\textbf{\Large Llama 3.2-11B}\\\textbf{\Large Vision-Instruct}} &
    \cellcolor{qwen2b!50}\textbf{\Large Qwen3-2B-Instruct} &
    \cellcolor{qwen4b!50}\textbf{\Large Qwen3-4B-Instruct} &
    \cellcolor{qwen8b!50}\textbf{\Large Qwen3-8B-Instruct} \\
    \midrule
    \rowcolor{navyblue!5}
    \multicolumn{10}{c}{\textcolor{black}{\Large\textbf{Scene Understanding}}} \\
    \textbf{\Large Q1} &
      \ci{97.2}{2.1}{2.8} &
      \ci{100.0}{0.0}{0.0} &
      \ci{100.0}{0.0}{0.0} &
      \ci{96.5}{2.8}{3.5} &
      \ci{95.8}{2.8}{3.5} &
      \ci{98.6}{1.4}{2.1} &
      \ci{34.0}{7.6}{7.6} &
      \ci{99.3}{0.7}{1.4} &
      \ci{97.2}{2.1}{2.8} \\
    \textbf{\Large Q2} &
      \ci{95.1}{3.5}{3.5} &
      \ci{100.0}{0.0}{0.0} &
      \ci{100.0}{0.0}{0.0} &
      \ci{94.4}{3.5}{4.2} &
      \ci{97.9}{2.1}{2.8} &
      \ci{95.8}{2.8}{3.5} &
      \ci{95.1}{3.5}{3.5} &
      \ci{100.0}{0.0}{0.0} &
      \ci{100.0}{0.0}{0.0} \\
    \textbf{\Large Q3} &
      \ci{100.0}{0.0}{0.0} &
      \ci{100.0}{0.0}{0.0} &
      \ci{100.0}{0.0}{0.0} &
      \ci{96.5}{2.8}{3.5} &
      \ci{95.8}{2.8}{3.5} &
      \ci{82.6}{5.6}{6.2} &
      \ci{100.0}{0.0}{0.0} &
      \ci{100.0}{0.0}{0.0} &
      \ci{100.0}{0.0}{0.0} \\
    \midrule
    \rowcolor{navyblue!5}
    \multicolumn{10}{c}{\textcolor{black}{\Large\textbf{Spatial Reasoning}}} \\
    \textbf{\Large Q4} &
      \ci{83.3}{5.6}{6.2} &
      \ci{98.6}{1.4}{1.7} &
      \ci{95.8}{2.8}{3.5} &
      \ci{79.9}{6.2}{6.9} &
      \ci{89.9}{4.5}{4.9} &
      \ci{84.8}{4.8}{5.2} &
      \ci{91.0}{4.2}{4.9} &
      \ci{97.2}{2.1}{2.8} &
      \ci{87.5}{4.9}{5.6} \\
    \textbf{\Large Q5} &
      \ci{41.7}{8.3}{8.3} &
      \ci{72.9}{6.9}{7.6} &
      \ci{64.6}{7.6}{7.6} &
      \ci{29.9}{6.9}{7.6} &
      \ci{55.6}{7.6}{8.3} &
      \ci{38.6}{7.6}{8.3} &
      \ci{36.1}{7.6}{8.3} &
      \ci{46.5}{8.3}{8.3} &
      \ci{47.2}{8.3}{8.3} \\
    \midrule
    \rowcolor{navyblue!5}
    \multicolumn{10}{c}{\textcolor{black}{\Large\textbf{Visual Perspective Taking}}} \\
    \textbf{\Large Q6} &
      \ci{48.6}{7.6}{8.3} &
      \cellcolor{bestclosed}\ci{87.5}{4.9}{5.6} &
      \ci{59.0}{7.6}{8.3} &
      \ci{38.9}{7.6}{8.3} &
      \ci{56.2}{8.3}{8.3} &
      \ci{49.3}{8.3}{8.3} &
      \ci{49.3}{8.3}{8.3} &
      \cellcolor{bestopen}\ci{54.2}{8.3}{8.3} &
      \ci{54.2}{7.6}{8.3} \\
    \textbf{\Large Q7} &
      \ci{30.2}{7.3}{7.3} &
      \cellcolor{bestclosed}\ci{59.0}{7.6}{8.0} &
      \ci{39.6}{7.6}{8.3} &
      \ci{34.0}{7.6}{7.6} &
      \ci{35.1}{6.9}{6.9} &
      \ci{31.9}{7.6}{7.6} &
      \ci{20.1}{6.2}{6.9} &
      \cellcolor{bestopen}\ci{37.5}{7.6}{7.6} &
      \ci{36.8}{7.6}{7.6} \\
    \bottomrule
    \ifdoublerule \specialrule{\heavyrulewidth}{0pt}{0pt}\fi
   \end{tabular}
   \end{adjustbox}
  }
\end{table*}

\paragraph{Models Evaluated.}
We tested nine popular models, including open source models -- Llama-3.2-11B-Vision-Instruct (11 December 2024)~\cite{meta2024llama}, Qwen3 (2B, 4B, and 8B)-Instruct~\cite{yang2025qwen3technicalreport} (1 November 2025) -- and five closed source models -- GPT-4-Turbo (9 April 2024), GPT-4o (6 August 2024)~\cite{openai2023gpt4v}, Claude 3 Sonnet (29 February 2024), Claude 3.5 Sonnet (20 June 2024)~\cite{anthropic2023claude} and Gemini Robotics-ER 1.5 (March 12, 2025)~\cite{geminiroboticsteam2025geminiroboticsbringingai}. Models' performance for each question is presented in Table~\ref{tab:model_comparison}. Figure~\ref{fig:main_fig} presents the performance averaged by categories. All error bars represent 95\% CIs calculated using bootstrapping (10,000 iterations).

\vspace{-10pt}
\paragraph{Random Classifier.}
Figure~\ref{fig:main_fig} also plots a random classifier baseline, i.e. the performance achieved by selecting answers uniformly at random from the permissible answer pool, for more details, see~\ref{app:random_baseline}.
\vspace{-10pt}
\paragraph{Scene Understanding.}
All nine models performed strongly on scene understanding tasks, reflecting their ability to recognize objects and count humanoid minifigures. GPT-4o and Gemini Robotics-ER 1.5 achieved perfect performance at $100.0\%~^{+0.0}_{-0.0}$ \metricLabel{}, closely followed by Qwen3-4B-Instruct ($99.8\%~^{+0.2}_{-0.5}$), Qwen3-8B-Instruct ($99.1\%~^{+1.2}_{-1.2}$), and GPT-4-Turbo ($97.5\%~^{+1.9}_{-2.1}$). Claude-3.5-Sonnet ($96.5\%~^{+2.5}_{-3.2}$), Claude~3 Sonnet ($95.8\%~^{+3.0}_{-3.7}$), and Llama-3.2-11B-Vision-Instruct ($92.4\%~^{+3.5}_{-3.9}$) also demonstrated high performance. Qwen3-2B-Instruct achieved $76.4\%~^{+3.7}_{-3.9}$ overall, with particularly low performance on Q1 ($34.7\%~^{+7.6}_{-7.6}$) driven by occasional misclassification of the white background as a non-humanoid minifigure or as a \textit{white brick}, though it recovered to near-perfect performance on Q2 and Q3. This suggests that identifying \textit{what} is in the scene -- in this instance, how many humanoid minifigures or objects are present, and whether they share the same surface -- has become a relatively routine task for modern VLMs, with smaller models occasionally struggling with certain visual ambiguities.

\begin{figure*}[ht!]
\centering
\includegraphics[width=\textwidth]{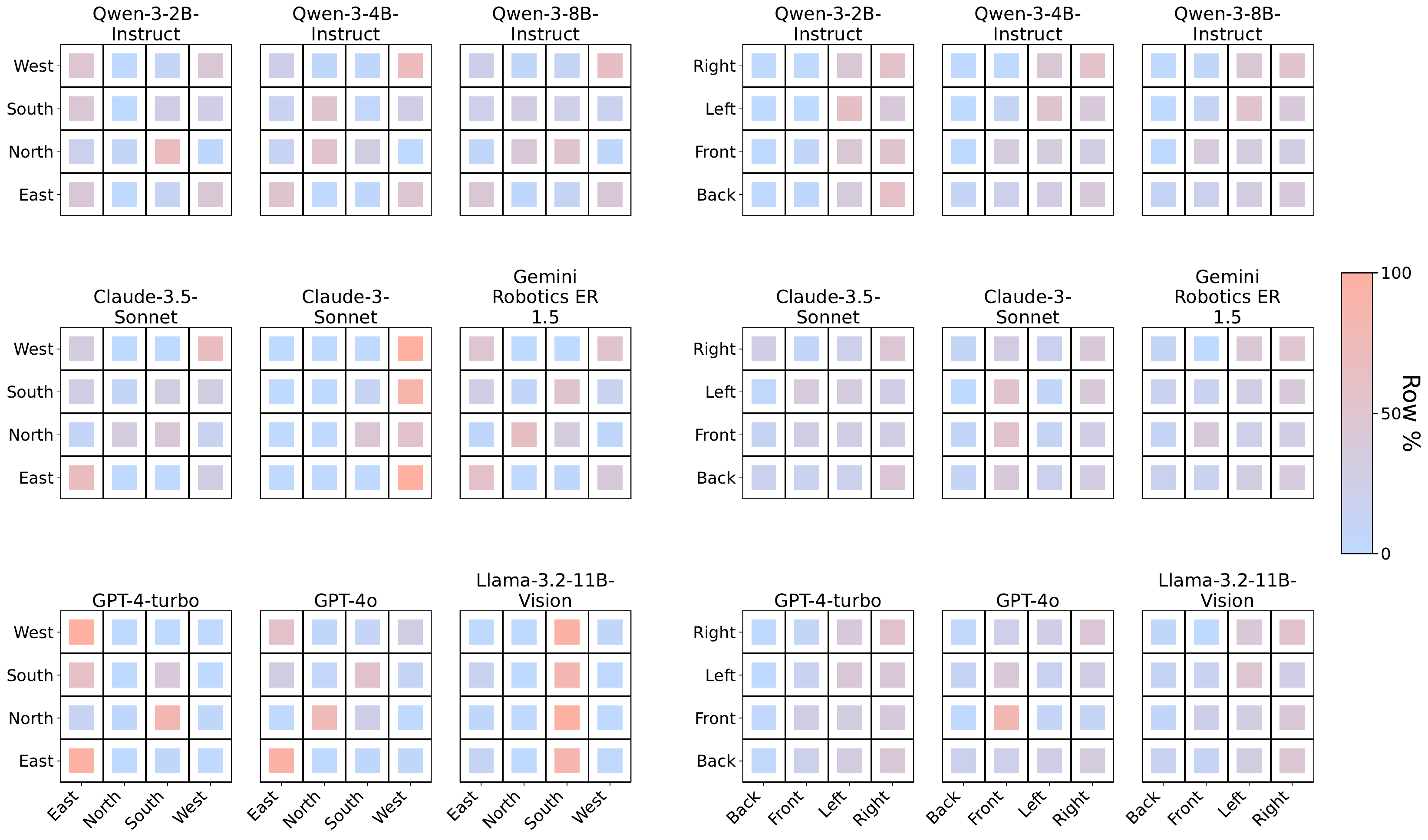}
\caption{\small Row-normalized co-occurrence heatmaps for nine models on Q5 (cardinal directions; \textbf{left}) and Q7 (egocentric directions; \textbf{right}). Q5 reveals strong \emph{directional bias}: GPT-4-Turbo collapses onto \emph{East} (with \emph{South} as a frequent fallback); GPT-4o strongly favors \emph{East} (and somewhat \emph{South}); Llama-3.2-11B-Vision concentrates on \emph{South}; and Claude-3-Sonnet almost always predicts \emph{West}. Qwen3-(2B/4B/8B)-Instruct systematically underuse \emph{North}, with 2B over-predicting \emph{East} and 4B particularly favoring \emph{West/North}. In contrast, Q7 exhibits milder egocentric asymmetries, with probability mass drifting toward \emph{Right/Left} (e.g., GPT-4-Turbo, Llama-3.2-11B-Vision) or \emph{Front/Right} (GPT-4o). Collectively, these patterns reveal a systematic preference for specific default bearings rather than balanced spatial reasoning.\vspace{-10pt}}
\label{fig:bias}
\end{figure*}

\vspace{-10pt}
\paragraph{Spatial Reasoning.}
Models fare considerably weaker on spatial reasoning tasks. Although they performed relatively well when localizing objects relative to the humanoid minifigure (Q4), the \metricLabel{} dropped significantly when the models had to determine the humanoid minifigure's facing direction (Q5). We hypothesize this discrepancy arises because Q4 involved an extrinsic reference frame, where objects are localized relative to the humanoid minifigure using the fixed cardinal directions established by the image orientation. Q5, however, required understanding an intrinsic reference frame based on the humanoid minifigure's own orientation, demanding that models interpret body posture cues to determine facing direction.

For Q5, we observed that GPT-4-Turbo, Claude 3 Sonnet, Qwen3 (2B, 4B, and 8B)-Instruct, and Llama-3.2-11B-Vision-Instruct were susceptible to \emph{directional bias}, see Figure~\ref{fig:bias}. Namely, they favored certain cardinal directions, for example, GPT-4-Turbo focused on east and south, completely omitting other directions. For this model, we ran additional detailed experiments in Section~\ref{sec:bias}. Namely, we systematically tested variations such as removing secondary objects, zooming in on humanoid minifigures, explicitly labeling cardinal directions (N, S, E, W) in the visual tasks, and even replacing humanoid minifigures with human faces. None of these was able to significantly mitigate the GPT-4-Turbo’s directional bias. This suggests that some models may rely on \textit{linguistic priors} or \textit{memorized defaults} rather than genuinely engaging in spatial reasoning.

\vspace{-10pt}
\paragraph{Visual Perspective Taking.}
This task assessed models on two VPT levels: determining if the humanoid minifigure saw an object (Q6; Level-1) and identifying the object's relative position from the humanoid minifigure's perspective (Q7; Level-2). Among open-source models, Qwen3-4B-Instruct achieved the highest performance, reaching $54.2\%^{+8.3}_{-8.3}$ on Q6, though this represented only a modest 4.2 percentage point improvement over the random baseline. In comparison, GPT-4o performed well in Q6 ($87.5\%^{+4.9}_{-5.6}$), making fewer errors than its peers. By contrast, Claude 3 Sonnet often failed Q6 by rejecting the question's premise -- insisting that a humanoid minifigure \textit{cannot really see} -- a recurring error pattern that occurred systematically in 39/144 instances. These issues, largely fixed in Claude 3.5 Sonnet, primarily involved rejecting the premise or misidentifying objects. For example, common premise rejection errors included statements like \textit{...inanimate LEGO toy, it does not possess actual vision...} Similarly, object misclassification occurred, such as when the model stated \textit{There is no dog visible; the black piece appears to be a weapon...} However, for Q7, all models had difficulties, frequently misclassifying objects located behind the humanoid minifigure as being to its left or right, as indicated on co-occurrence matrices shown in Figure~\ref{fig:bias}.

This discrepancy between detecting scene content and simulating the humanoid minifigure’s true perspective highlights a critical shortfall in current VLMs. Recognizing objects does not necessarily equate to robust geometric reasoning or an inferential grasp of spatial relationships -- cognitive skills in humans linked to mental rotation and perspective taking. Understanding these shortcomings requires further studies. In this work, we performed a simple analysis starting from the observation that both Q6 and Q7 can be perceived as a combination of Q4 and Q5 and very simple reasoning (e.g. for Q6, the answer is positive iff the object is located in the same direction as the humanoid minifigure is facing, these are determined answering Q4 and Q5 resp.). Following that, one could hypothesize that poor performance on Q5 is the root cause of problems with Q6 and Q7. To test it, we added the ground truth Q5 answer (i.e. the humanoid minifigure's facing direction) to the prompt for Q6. It turned out that this results in modest improvements, suggesting that the aforementioned decomposition is not entirely accurate. We provide more discussion in \ref{app:hint}.

\subsection{Orientation vs Perspective Test}\label{app:hint}
We started by looking at how models handle questions about what a humanoid minifigure in an image might be seeing (Q6). One plausible route to answering such questions is to first infer which direction the humanoid minifigure is facing (Q5). For example, if an object is to the north and the humanoid minifigure is facing north, then the humanoid minifigure likely sees the object. This led us to ask: \textit{are the difficulties models have with visual perspective taking (Q6) mainly driven by problems identifying the humanoid minifigure's facing direction (Q5)?}

\begin{figure}[ht!] 
  \centering
  \includegraphics[width=\linewidth]{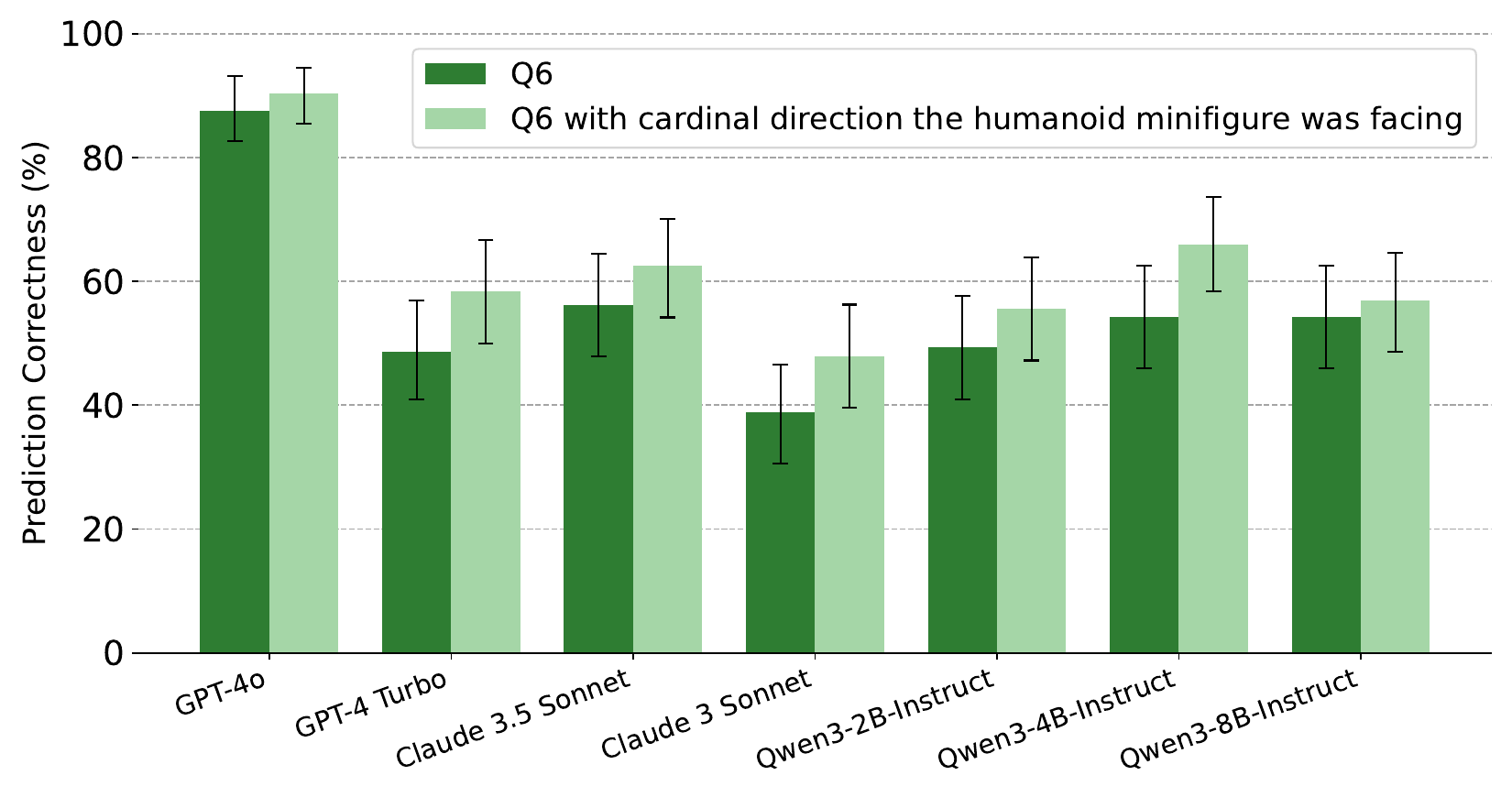}
  \caption{\small Comparison of VLM performance on visual perspective taking (Q6) with and without an explicit orientation hint (gold-standard Q5 answer), showing only marginal \metricLabel{} improvement.}
\label{fig:q6hint}
\end{figure}

To investigate this, we ran an experiment across 144 visual tasks. We specifically tested the VLMs on Q6, but with a helpful modification to the prompt: we explicitly told the model which cardinal direction the humanoid minifigure was facing, using the gold-standard answer from Q5 as a hint. Our reasoning was that if the struggle with Q5 was the primary bottleneck for Q6, providing this directional information should lead to a significant performance increase. However, as shown in Figure \ref{fig:q6hint}, adding this hint resulted in only marginal improvements. This suggests that the models' difficulties with visual perspective taking might be more complex than simply getting the orientation wrong, and solving Q5 alone does not automatically lead to correctly answering Q6.

\subsection{Directional Bias}\label{sec:bias}
In the results section, we described how models struggle with spatial judgments, particularly when determining the facing direction of a humanoid minifigure in Q5. We observed that some VLMs, such as GPT-4-Turbo, suffer from \emph{directional bias}, e.g., favoring directions like east or south. We aim to determine the source of the observed directional biases: do they originate primarily from challenges in visual perception, from how the prompt language is interpreted, or inherently biased, or from more fundamental issues within the model's spatial reasoning capabilities?

To this end, we are systematically investigating how specific interventions influence their judgments. We explored two paths: \textit{modifying the visual input} and \textit{manipulating the textual prompts}.

\paragraph{Removing Objects.}
\begin{figure}[ht!]
  \centering
  \includegraphics[width=\linewidth]{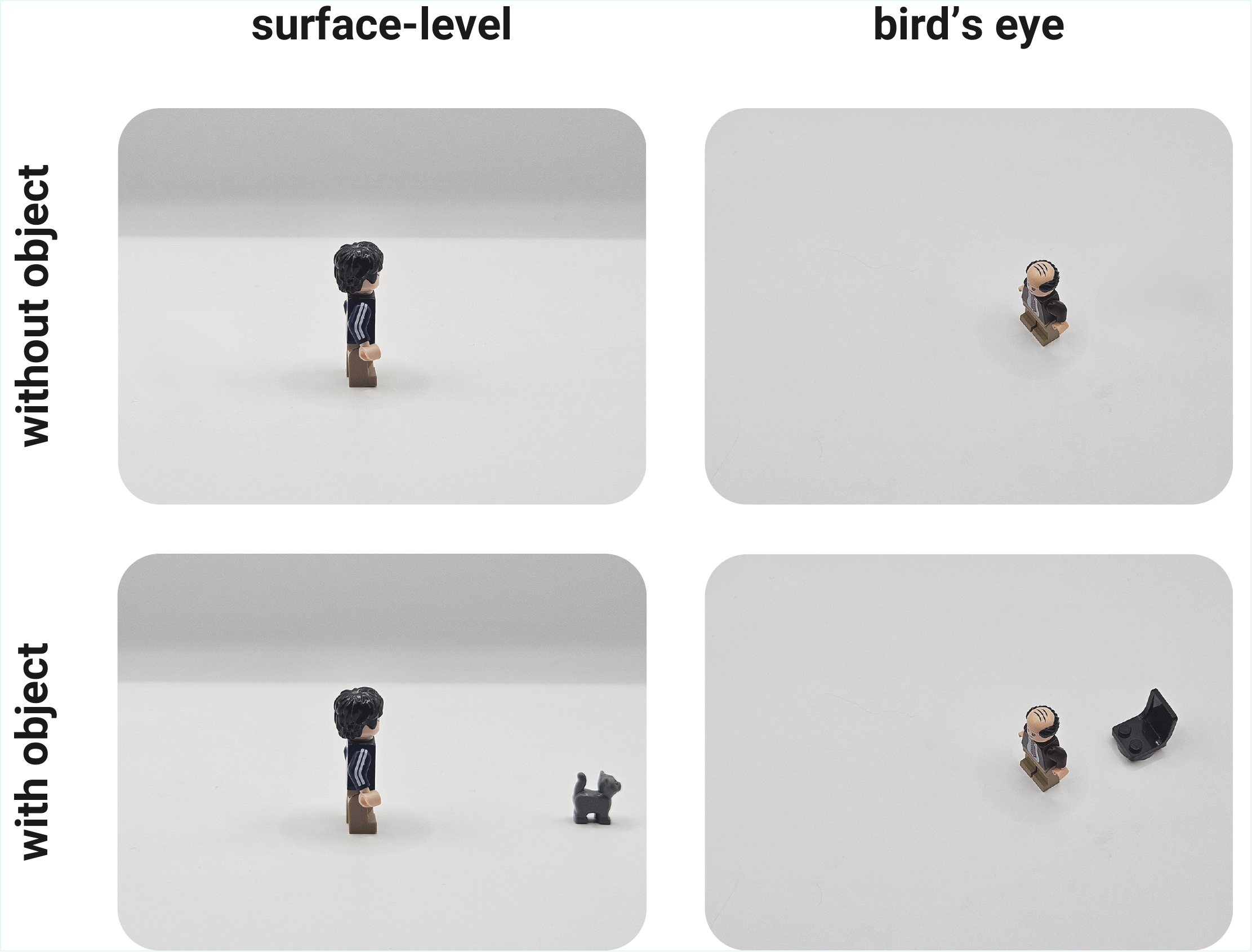} 
  \caption{\small Illustration of the object removal process, used in investigating if the presence of contextual objects impacts the model's orientation predictions (Q5). Top: visual tasks with objects. Bottom: corresponding visual tasks without objects. Left/Right: surface-level/bird's-eye.}
\label{fig:no_object}
\end{figure}
We hypothesized that extraneous objects might contribute to GPT-4-Turbo's directional bias. To test this, we removed items like cats, chairs, etc., from the test images, see Figure~\ref{fig:no_object}, and re-ran the 36 visual tasks assessing humanoid minifigure orientation (e.g., \textit{surface-towards-left}, \textit{birds-eye-towards-right}). 
 Although removing objects caused pointwise changes in 5 predictions and slightly increased the frequency of \textit{south} answers (from 1 to 5), \textit{east} remained the vastly predominant response (31 times). Therefore, object removal did not substantially reduce the model's tendency towards \textit{east}, indicating this bias is robust and not merely an artifact of contextual items.

\paragraph{Zoom.}
\begin{figure}[ht!]
  \centering
  \includegraphics[width=\linewidth]{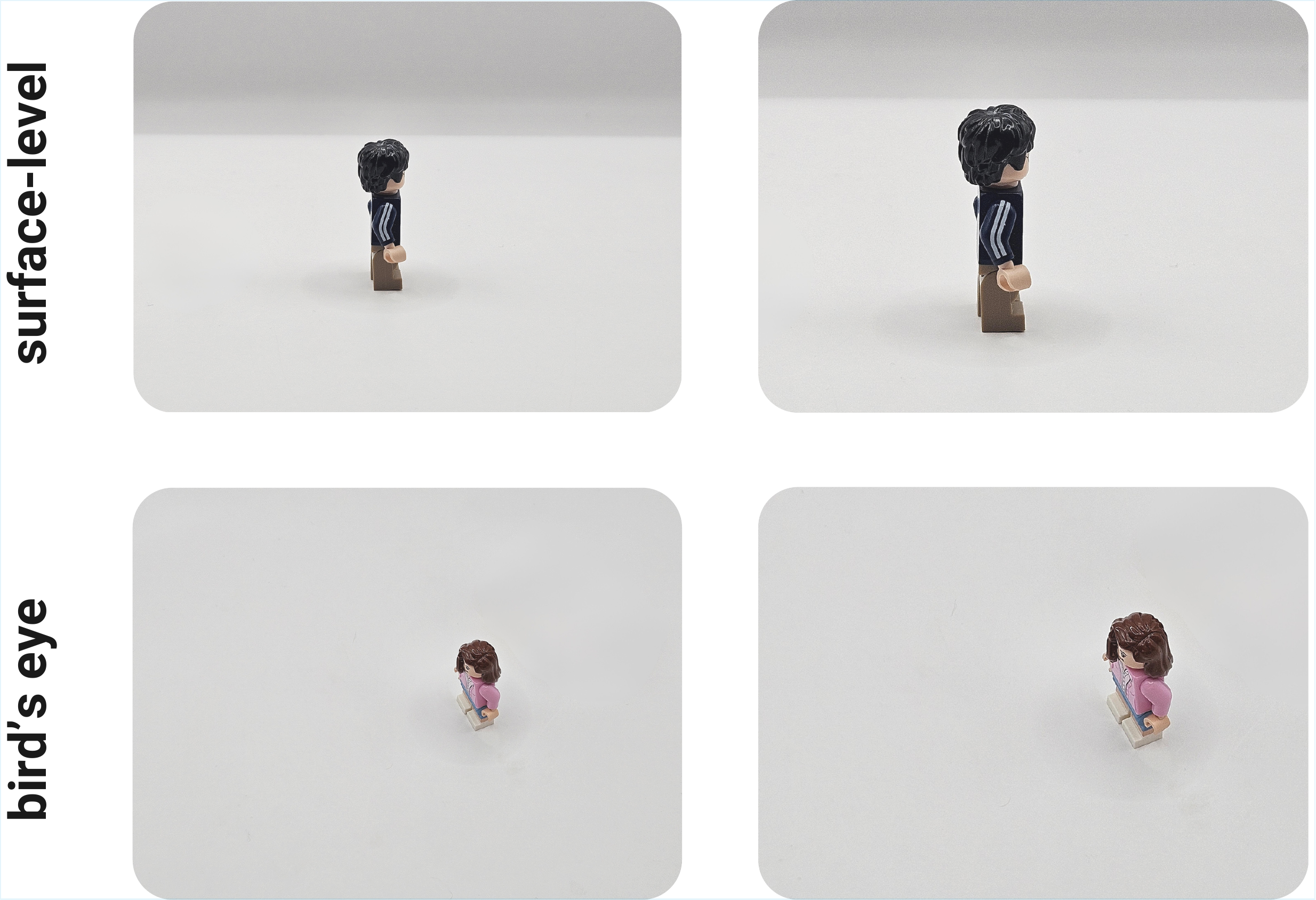} 
  \caption{\small Examples of humanoid minifigure orientation tasks (Q5) at different zoom levels. Left to right: original image with 10\% zoom, 30\% zoom, and 50\% zoom.}
\label{fig:zoom}
\end{figure}
To test if the bias was related to perceiving details, we conducted an experiment gradually zooming in (10\%, 30\%, 50\%) on the visual task without object, see Figure~\ref{fig:zoom}. Performance on the 36 tasks remained poor, with \metricLabel{} consistently around {\ $44.4\%^{+16.7}_{-16.7}$} (e.g., {\ $44.4\%^{+16.7}_{-16.7}$}/{\ $41.7\%^{+16.7}_{-16.7}$}/{\ $47.2\%^{+16.7}_{-16.7}$} for 10\%/30\%/50\% zoom). Importantly, this low accuracy was accompanied by the same persistent directional bias: GPT-4-Turbo continued to predominantly output east, irrespective of zoom level. This persistence, even when orientation details were magnified, suggests the bias is not merely a perceptual limitation regarding fine details, but potentially points to deeper issues in the model's spatial reasoning.

\paragraph{On-Visual Task Cardinal Hints.}\label{app:on_the_image}

\begin{figure}[ht!] 
  \centering
  \includegraphics[width=\linewidth]{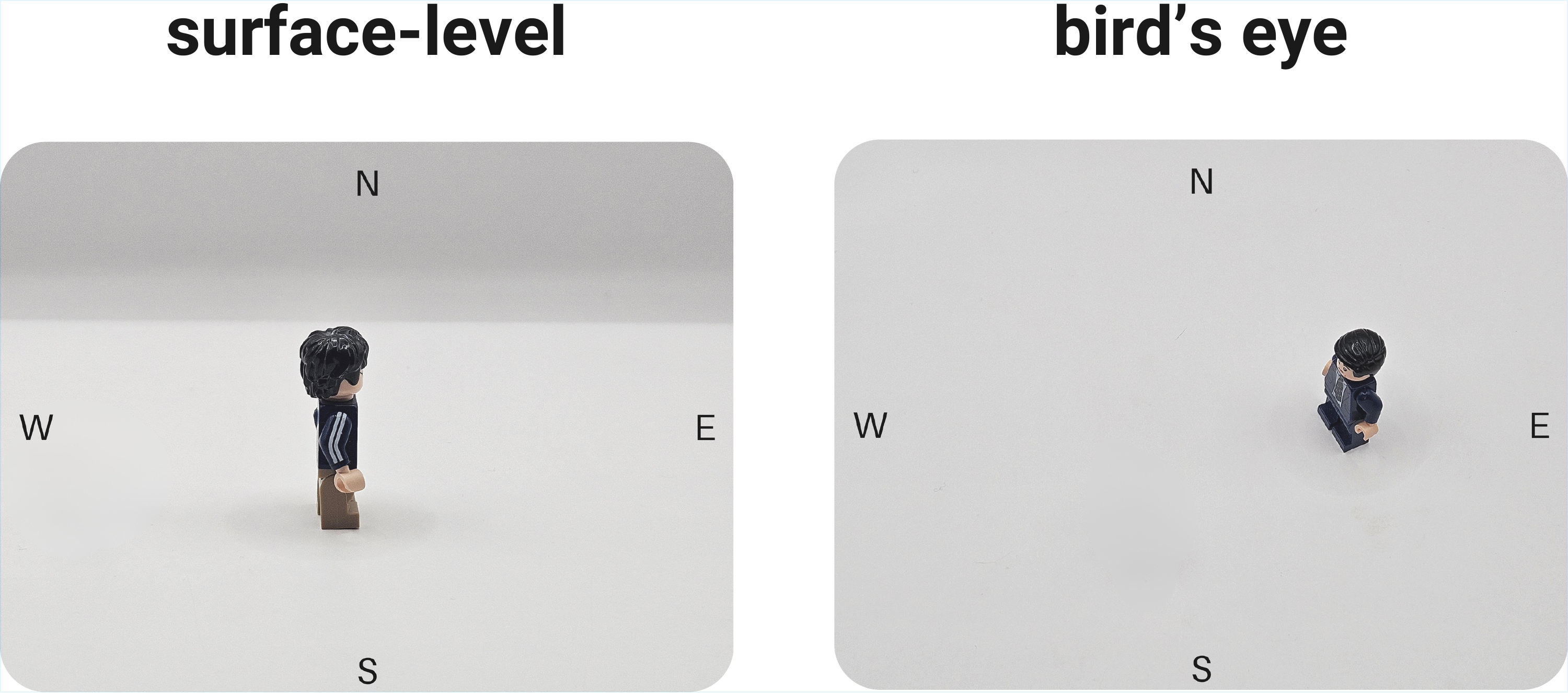} 
  \caption{\small Surface-level (left) and bird's-eye (right) views of the visual task, showing \textit{N S E W} marks on the image used for the on-visual task cardinal hints experiment. \vspace{-10pt}}
\label{fig:cardinal_overlay}
\end{figure}

To isolate reference-frame ambiguity, we overlaid \textit{N E S W} markers on each visual task, as illustrated in Figure~\ref{fig:cardinal_overlay}. Even with these cues, \metricLabel{} remained low at {\ $34.3\%^{+14.3}_{-17.1}$}, and the model still selected \textit{east} in 27 of the 36 trials. Because every stimulus explicitly provided both the scene geometry and its coordinate system, any residual error could have arisen from the model’s internal mapping between visual layouts and directional tokens rather than from mis-perceiving \textit{north}. In this light, the persisting bias may reflect a hard-wired prior embedded somewhere in the model’s spatial-reasoning pipeline.
\begin{figure}[ht!]
  \centering
  \includegraphics[width=\linewidth]{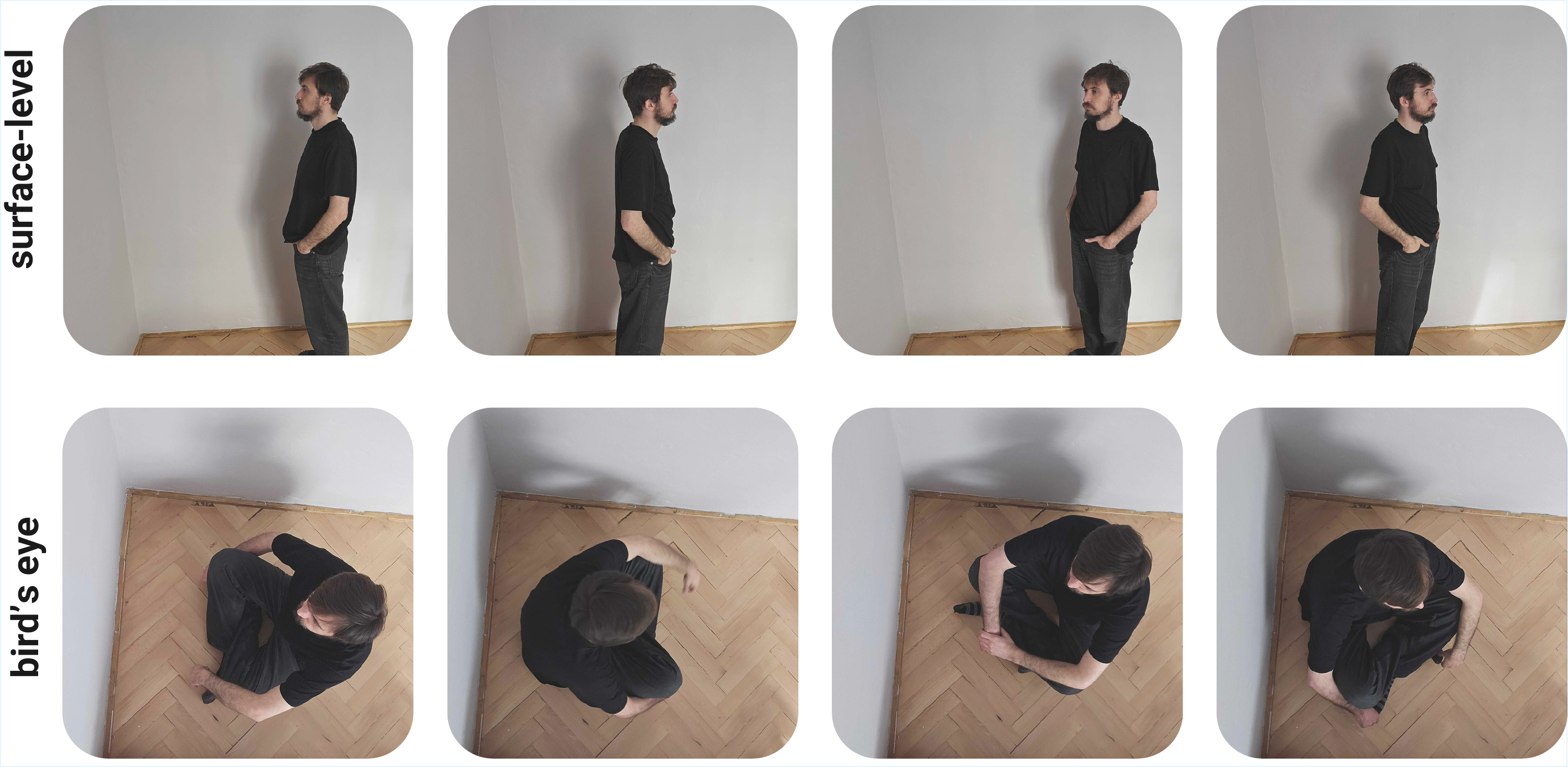}
  \caption{\small Surface-level (top row) and bird's-eye (bottom row) views of a real person used in the influence of subject type experiment.}
\label{fig:human_test}
\end{figure}

\vspace{-10pt}
\paragraph{Influence of Subject Type.}
To investigate if the observed directional bias was specific to the subject type (i.e., a plastic humanoid minifigure versus a real person), we used 8 images of a real person, consisting of 4 surface-level and 4 bird's-eye view shots, as shown in Figure~\ref{fig:human_test}. Accordingly, we modified the prompt to ask about the direction the \textit{person} was facing, replacing \textit{humanoid minifigure}. Notably, for all 8 images, GPT-4-Turbo responded that the person was facing east (e.g., answering \textit{The person in the image is facing east}). This unanimous east prediction mirrored the strong bias observed with the humanoid minifigure images. Consequently, this persistence suggests that the directional bias is not solely attributable to the specific nature or structure of the humanoid minifigure itself, but may stem from a more general aspect of the model's processing.

\section{Discussion} Our synthetic setting represents a lower bound for VLM spatial reasoning failures: the controlled environment provides perfect lighting, no occlusions, and distinct object separation. That VLMs fail under these optimal conditions suggests deficits fundamental to their spatial reasoning capabilities, not artifacts of perceptual noise in natural datasets (e.g., ScanNet~\cite{dai2017scannet}). The persistent directional biases and failure of multiple interventions—including explicit orientation cues, zooming, cardinal markers, and real human subjects—indicate models cannot compute perspective-dependent spatial relations, instead relying on linguistic priors over visual facts. We uncover a systematic mismatch between model predictions and ground truth on VPT tasks, in which models privilege linguistic priors over visual information; this may have implications for downstream alignment. Addressing these limitations requires architectural innovations that properly ground spatial reasoning in visual perception rather than learned textual associations. These failures highlight potential risks for applications that depend critically on reliable perspective-taking, such as autonomous driving and robotic assistance.
\section{Conclusions}
Our psychology-grounded diagnostic reveals a critical gap between VLMs' strong scene understanding capabilities and their markedly weaker performance on spatial reasoning and visual perspective taking. While models achieved near-perfect accuracy identifying objects and humanoid minifigures, their performance degraded substantially when determining spatial relationships and perspective-dependent locations. Future work should focus on architectures incorporating explicit geometric representations, specialized training protocols emphasizing mental rotation, and hybrid approaches combining symbolic spatial reasoning with learned representations to bridge the gap between recognition and true spatial understanding.
\section{Limitations}
Our diagnostic focuses on simple spatial configurations with single humanoid minifigures and objects in static images, which may not fully capture the complexity of real-world perspective taking involving multiple agents, dynamic scenarios, and complex occlusions. The limited spatial coverage (four cardinal positions, two orientations) and reliance on LEGO minifigures, while enabling experimental control, restricts generalization to more diverse and naturalistic settings. Additionally, we evaluated nine models in their out-of-the-box configurations without extensive prompt engineering, and our interventions were primarily conducted on GPT-4-Turbo, leaving room for exploring alternative reasoning strategies and newer model architectures.

{
    \small
    \bibliographystyle{ieeenat_fullname}
    \bibliography{main}
}

\clearpage
\setcounter{page}{1}
\maketitlesupplementary

\appendix

\section{Data}

\subsection{Gold-Standard Answer}\label{app:gold_dist}
In Table~\ref{tab:golddist} we list the gold-standard answers distribution.

\begin{table}[h]
\caption{Gold–standard answer distribution for each question type (Q1–Q7).} 
\centering
\footnotesize
\begin{tabularx}{\columnwidth}{lX}
\toprule
\textbf{Q} & \textbf{Gold-Standard Answers (\% of items)} \\
\midrule
Q1 & 1 (100.0) \\
Q2 & 1 (100.0) \\
Q3 & yes (100.0) \\
Q4 & east (15.3), east, north (10.4), east, south (8.3), north (14.6), north, west (10.4), south (18.1), south, west (6.9), west (16.0) \\
Q5 & east (15.3), east, south (9.7), north (25.0), south (25.0), south, west (7.6), west (17.4) \\
Q6 & no (50.0), yes (50.0) \\
Q7 & back (31.9), back, left (4.9), back, right (13.2), front (39.6), front, left (7.6), front, right (2.8) \\
\bottomrule
\end{tabularx}
\label{tab:golddist} 
\end{table}
\subsection{\metricLabelcapital{}} \label{app:acc}

\begin{figure*}[ht!]
\centering
\includegraphics[width=\textwidth]{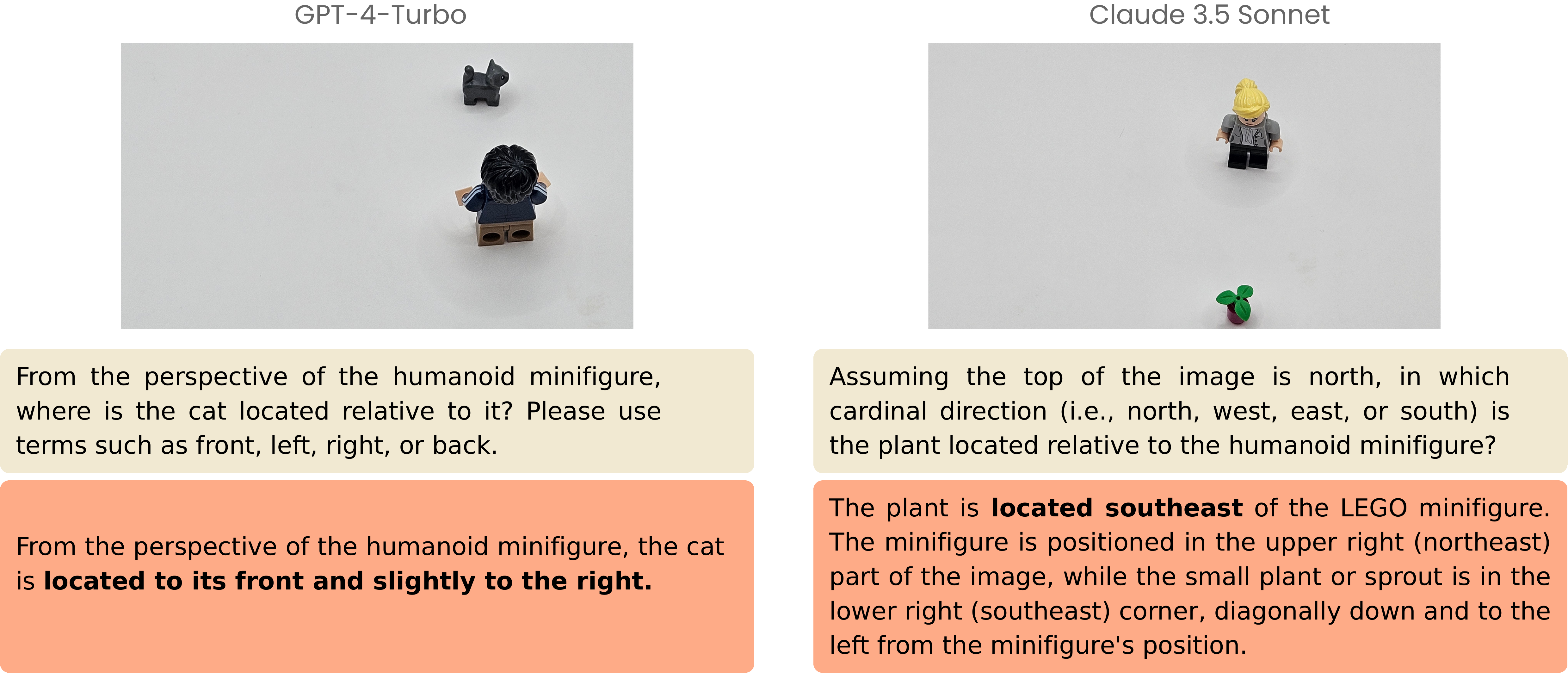}
\caption{\small  Example model responses to Q5 and Q7 questions. Left: GPT-4 Turbo answers \textit{front and slightly to the right} (gold-standard answer: front). Right: Claude 3.5 Sonnet answers \textit{southeast} (gold-standard answer: south).}
\label{fig:errors}
\end{figure*}

Models sometimes generated compound answers -- for instance, \textit{northeast} in Q4 and Q5, or \textit{back and slightly to the left} in Q7 (see Figure~\ref{fig:errors} and Table~\ref{tab:answer_distribution}). Because these responses contained multiple components, our evaluation needed to acknowledge partial as well as fully correct answers.

To tackle this, we employed a precision-based metric that rewarded models for each correctly identified component while tolerating omissions. Assume that we evaluate $R$ responses for a given diagnostic questions and a model, like in Table \ref{tab:model_comparison}. The score $P$ reported in the table is called \emph{\metricLabel{}} and is defined as the mean precision across all $R$ responses:
$$P=\frac{1}{R}\sum_{i=1}^{R} P_i, \quad \text{where }P_i = P(M_i, G_i) = \frac{|M_i \cap G_i|}{|M_i|},$$
where $M_i$ (resp. $G_i$) is the set of components in the model's prediction (resp. gold-standard answer) for the $i$ response. 

The value of $P$ ranges from 0  to 1. We note that for the questions with single answers, like e.g., Q6, this metric is equivalent to standard accuracy. Moreover, we experimented with several other metrics that take into account partial correctness (e.g., the Jaccard index), and they all yielded similar results.

\subsection{Random Baseline} \label{app:random_baseline}
\begin{table}[h!]
\caption{\small Frequency of answer types for Q4, Q5, and Q7 across models.  
\emph{Comp.} (compound) denotes compound answers (e.g., \textit{northwest} to Q5);  
\emph{Sing.} (Single) denotes single-word answers (e.g., \textit{back} to Q7);  
\emph{Disc.} (Disclaimer) marks instances in which the model failed to provide a relevant answer (e.g., claiming no object is present when one is).
}
    \centering
    \footnotesize
    \setlength{\tabcolsep}{4pt}
    \begin{tabular}{@{}lccc@{}}
    \toprule
    \textbf{Model} & \textbf{Comp.} & \textbf{Sing.} & \textbf{Disc.} \\
    \midrule
    \multicolumn{4}{@{}l}{\textit{Question 4}} \\
    \midrule
    GPT-4-o                    & 21 & 123 & 0 \\
    GPT-4-Turbo                & 0  & 144 & 0 \\
    Claude 3.5 Sonnet          & 21 & 117 & 6 \\
    Claude 3 Sonnet            & 0  & 127 & 17 \\
    Llama-3.2-11B-Vision       & 33 & 111 & 0 \\
    Qwen3-2B-Instruct          & 0  & 144 & 0 \\
    Qwen3-4B-Instruct          & 0  & 144 & 0 \\
    Qwen3-8B-Instruct          & 0  & 144 & 0 \\
    Gemini Robotics ER 1.5     & 0  & 144 & 0 \\
    \midrule
    \multicolumn{4}{@{}l}{\textit{Question 5}} \\
    \midrule
    GPT-4-o                    & 0 & 144 & 0 \\
    GPT-4-Turbo                & 0 & 144 & 0 \\
    Claude 3.5 Sonnet          & 0 & 144 & 0 \\
    Claude 3 Sonnet            & 0 & 144 & 0 \\
    Llama-3.2-11B-Vision       & 1 & 143 & 0 \\
    Qwen3-2B-Instruct          & 0 & 144 & 0 \\
    Qwen3-4B-Instruct          & 0 & 144 & 0 \\
    Qwen3-8B-Instruct          & 0 & 144 & 0 \\
    Gemini Robotics ER 1.5     & 0 & 144 & 0 \\
    \midrule
    \multicolumn{4}{@{}l}{\textit{Question 7}} \\
    \midrule
    GPT-4-o                    & 10 & 134 & 0 \\
    GPT-4-Turbo                & 16 & 127 & 1 \\
    Claude 3.5 Sonnet          & 68 & 76  & 0 \\
    Claude 3 Sonnet            & 1  & 121 & 22 \\
    Llama-3.2-11B-Vision       & 2  & 139 & 3 \\
    Qwen3-2B-Instruct          & 0  & 144 & 0 \\
    Qwen3-4B-Instruct          & 0  & 144 & 0 \\
    Qwen3-8B-Instruct          & 0  & 144 & 0 \\
    Gemini Robotics ER 1.5     & 0  & 144 & 0 \\
    \bottomrule
    \end{tabular}
\label{tab:answer_distribution}
\end{table}
To establish a chance performance benchmark, we defined a random baseline classifier selecting one answer uniformly at random from the $N_q$ distinct choices available for question type $q$. For a given instance of question type $q$, the probability of this random choice being evaluated as correct is $\mathrm{Prob}(\text{success}_q) = \frac{k_q}{N_q}$, where $k_q$ is the number ($k_q \ge 1$) of acceptable answers specified in the gold-standard answers for that instance out of $N_q$ total options. This formulation accurately quantifies the expected success rate of uninformed random guessing under our evaluation protocol, accommodating multiple correct answers, based on the empirical distribution of $k_q$ values observed in our gold-standard answers dataset (detailed in~\ref{app:gold_dist}, Table~\ref{tab:golddist}). For instance, for Q4 ($N_4=4$), where 64\% of instances have $k_4=1$ ($\mathrm{Prob}=\frac{1}{4}$) and 36\% have $k_4=2$ ($\mathrm{Prob}=\frac{1}{2}$), the weighted average yields a chance level for this question type of $(0.64 \times \frac{1}{4}) + (0.36 \times \frac{1}{2}) = 0.16 + 0.18 = 0.34$.

Category-level chance performance was computed by averaging the chance levels of the constituent question types. For example, the scene understanding category comprises Q1 (chance $=\frac{1}{3}$), Q2 (chance $= \frac{1}{3}$), and Q3 (chance $= \frac{1}{2}$), resulting in an average category chance level of $(\frac{1}{3} + \frac{1}{3} + \frac{1}{2}) / 3 = \frac{7}{18} \approx 0.389$. Following this methodology across all categories yields the following random baseline classifier levels: scene understanding ($0.389$), spatial reasoning ($0.317$), and visual perspective taking ($0.411$).

\subsection{Co-occurrence  Matrix}\label{app:matrix}
Our co-occurrence matrices presented in the result section show how the model’s predictions line up with each gold-standard answer. To build it, we take all questions whose gold-standard answers include a particular label -- for example, \textit{north} in Q5, and within that subset simply count how often the model produced \textit{north}, \textit{east}, \textit{south}, or \textit{west}. Those four counts become the row for \textit{north}. Because a single question can have several gold answers and the model may mention several answers at once (such as \textit{northeast} in Q5 or \textit{back and slightly to the left} in Q7), one question can be counted in more than one row or column, so the values in a row may exceed the total number of questions. When every question has exactly one gold label and the model also outputs exactly one label, this co-occurrence table collapses to the ordinary single-label confusion matrix, with each row summing to the number of items.

\end{document}